# Developing an Informal-Formal Persian Corpus

**Vahide Tajalli**[1], **Fateme Kalantari**[2], **Mehrnoush Shamsfard**[3]
NLP Research Lab,
Faculty of Computer Science and Engineering,
Shahid Beheshti University, Tehran, Iran

**Abstract**

Informal language is a style of spoken or written language frequently used in casual conversations, social media, weblogs, emails and text messages. In informal writing, the language faces some lexical and/or syntactic changes varying among different languages. Persian is one of the languages with many differences between its formal and informal styles of writing, thus developing informal language processing tools for this language seems necessary. Such a converter needs a large aligned parallel corpus of colloquial-formal sentences which can be useful for linguists to extract a regulated grammar and orthography for colloquial Persian as is done for the formal language.

In this paper we explain our methodology in building a parallel corpus of 50,000 sentence pairs with alignments in the word/phrase level. The sentences were attempted to cover almost all kinds of lexical and syntactic changes between informal and formal Persian, therefore both methods of exploring and collecting from the different resources of informal scripts and following the phonological and morphological patterns of changes were applied to find as much instances as possible. The resulting corpus has about 530,000 alignments and a dictionary containing 49,397 word and phrase pairs.

**Keywords:** Persian, Informal Writing, Corpus, Colloquial Language.

## 1. Introduction

Informal language is more common when we speak. However, there are times when writing can be very informal, for instance, in weblog posts, social media comments, and text messages. Informal writing is in fact a reflection of linguistic features of colloquial speech in our written materials.

Informal Persian is different from its formal form both lexically and syntactically. A large amount of colloquial Persian data is created every day in the cyberspace and the media, thus developing informal language processing tools for this language seems necessary. Forming a Persian informal-formal parallel corpus will enable computer engineers and computational linguists to develop tools for converting these two styles automatically or process texts in both styles with a strong performance.

1 - vtajalli@ut.ac.ir
2 - fatemekalantari@hafez.shirazu.ac.ir
3 - m-shams@sbu.ac.ir

There are several studies on Persian informal language. Most of them have tried to suggest a uniform orthography for informal language. Tabibzadeh (2020), among all, has conducted a seminal work which reviews 112 Persian novels and dramas written over 100 years. He chooses 1697 informal words randomly from these works and based on these words, he categorizes and explains the features of informal Persian. Since all his data comes from the books they have partly approved forms by the authors and editors. However, the situation is different in the virtual space where the people break the linguistic norms and try to show their feelings through the words by creating new forms.

Moreover, there are some researches on converting Persian colloquial texts into formal ones. Armin and Shamsfard (2011) and Naemi et al. (2021) propose rule-based systems which only cover a small part of the data. In addition, they just handle the lexical changes and syntactic ones are left.

Rasooli et al. (2020) suggest an automatic method for standardizing colloquial Persian text. Their core idea is training a sequence-to-sequence translation model translating colloquial Persian to standard Persian. They have annotated a publicly available evaluation data consisting of 1912 sentences.

Abdi Khojasteh et al. (2020) propose a dataset for Large-Scale Colloquial Persian (LSCP) containing about 120M sentences from twitter for machine translation with universal and treebank-specific POS tags with dependency relations and translations in five languages. In order to annotate the datasets, they adopt a semiautomatic crowd-sourcing method.

Kabiri et al. (2022) develop an Informal Persian Universal Dependency Treebank (iPerUDT) with a total of 3000 sentences from Persian blogs and mention a few differences between formal and informal Persian.

Although LSCP and iPerUDT can be used to study the colloquial Persian in lexical and syntax levels, they are not parallel corpora and have no formal counterparts for informal data, therefore they cannot be directly used for inter-style conversions.

As is noticeable, the available resources and tools are insufficient for covering all aspects of this issue either due to applying rule-based methods and having limited rules or due to using data-driven methods with limited or incomplete data. Therefore, a converter with a big dataset which can transform informal into formal language in both lexical and syntactic levels is needed to fill this gap. This article is a report of an attempt to build this dataset. Moreover, the differences between formal and informal Persian writing styles will be reported in details. We are not going to propose a standard orthography for informal Persian, however, studying these differences and making parallel corpus of these two language styles help linguists with developing uniform and regulated grammar and orthography for informal Persian.

The article is organized as follows: the next section briefly introduces Persian language and its informal style. Section 3 explains the procedure of building this informal dataset. Section 4 explains the differences between formal and informal Persian. Section 5 represents the results and in the end, section 6 concludes the paper with pointing out the conclusions and further works.

## 2. Informal vs. Formal Persian

Persian is a pro-drop language with canonical Subject-Object-Verb word order which is written in Arabic script with some small adjustments. In this script some letters are written connected to their adjacent ones and short vowels do not normally appear in writing. Persian informal language is different from formal in many ways. In order to build a comprehensive corpus covering syntactic and lexical dimensions, we need to know the characteristics of Persian informal language and its writing style.

Informal writing style has some general characteristics including making use of interjections, more idiomatic and conversational expressions, contractions, and imprecise words. Moreover, sentences are shorter since appositive phrases and complicated structures are not normally used in the informal language, whereas both fragments and run-on sentences are acceptable.

People break some rules of standard writing style and devise different writing methods to be able to convey the tone along with the meaning as far as possible (Tartar, 2018). Most of these changes are made in line with the principle of language economy and include the phonetic, lexical, syntactic and semantic aspects of the language (Eftekhari, 2014).

Apart from the fact that informal Persian is associated with particular choices of grammar and vocabulary, there are many formal words and expressions changing in informal language. Persian informal writing style is called *shekaste-nevisi* literally translated as "broken-writing", indicating that many formal words are cut down in informal Persian. In the present study, typical informal language used by Iranians has been considered and its informal writing style has been investigated in detail to develop the dataset.

## 3. Developing the Dataset

In this section we discuss our methodology in extracting candidate sentences, choosing appropriate ones, transforming them into formal sentences and making the alignments.

### 3.1 Extracting Informal Sentences from Available Resources

Sentences could be either selected from external sources or generated by the data linguists. In order for the linguistics teams to have access to a great variety of sources, they were provided with texts derived from online crawling of social networks, websites and blogs as well as some scripts of books, screenplays and movie subtitles. Before distributing the sources among team members, fonts were standardized and texts were normalized as far as possible.

There were other sources including different messengers and everyday conversations that could be considered by the linguistics teams. Since the study aimed to cover all styles of writing, we attempted to use every sources reasonably, depending on the level of usage. Table 1 shows the distribution of external sources and the number of informal sentences extracted from each one.

*Table 1. Sources of informal sentences and their distributions*

| Source of data | # of extracted sentences | description |
|---|---|---|
| Instagram | 9,625 | sentences with 26-40 words length including at least four informal words |
| Twitter | 7,000 | informal posts with corona-related content |
| Web Pages | 293,426 | extracted from more than ten websites and forums |

| Weblogs | 26,146 | extracted from three popular weblogs |
|---|---|---|
| Books | 124,130 | extracted from nine e-books and collections of short stories |
| Movies | 179,290 | extracted from six movie subtitles and play scripts |
| Total | **639,617** | |

In order to extract data, pages were crawled and sentences with the length of 26-40 tokens (space separated) including at least 4 informal words were selected. As a result, about 640,000 informal sentences were provided to the linguistics teams for searching the proper data. Finally, 50,000 sentences were selected or generated and entered into the dataset. More than 50% of them were reviewed and corrected or confirmed by two linguist leaders.

### 3.2 Software Tool for data gathering and preparation

Aiming to create the dataset, a software tool was developed letting the users enter data records. Each record included an informal sentence, its formal equivalent and their alignments in word and phrase levels. For each record, time and date of data entry, the data provider and the source of the informal sentence were saved and were searchable.

In order to speed up the development process, the system employed some automatic methods for suggesting the alignments using the previous found alignments, according to their frequency of occurrence and the context of the aligned word. The annotators checked the system's alignment suggestion to accept or correct it.

The tool managed data entry, data revision and confirmation, report generation, accounting, upload and download of raw and annotated corpus and some automatic data processing tasks for data verification and generation. For example, normalizing input sentences, checking for missing or inconsistent alignments and suggesting alignments were among automatic data processing tasks of the developed software. Fig. 1 shows a screenshot of data entry in this tool.

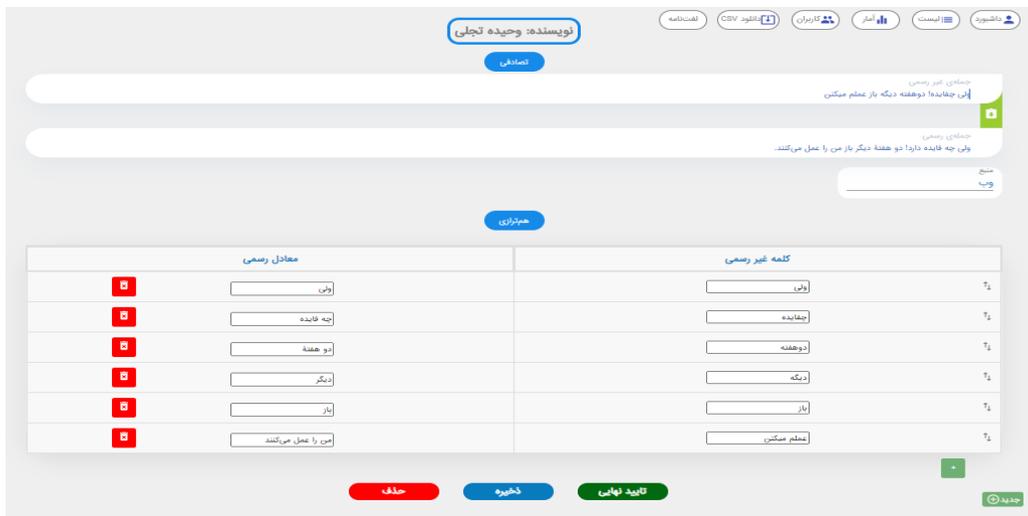

*Figure 1. An entry of dataset in the data gathering software*

### 3.3  Data Entry

Exploring the resources and spotting the linguistic points, we began to highlight the features of Persian informal language. 50,000 pairs of formal-informal sentences with specified alignments were supposed to be entered into the dataset. In order to decide the formal alignment, minimum changes were made and paraphrasing was not applied. Slang words and phrases were not replaced. There were a few expressions and utterances with no near formal equivalents; for these cases a negotiated equivalent was chosen. Formal sentences were entered with correct punctuations.

The style of writings seemed mostly to be affected by age, education level, and social group membership of language users. We attempted to cover all the levels as far as possible. As previously mentioned, many Persian words including the largest number of verbs have an abridged informal form. They were all replaced by formal word forms.

Rare mistakes like uncommon spelling mistakes in informal sentences were edited before entering but common mistakes were kept and edited in the formal equivalents. Some common spelling mistakes are the result of having more than one character for a sound in the Persian alphabet. The frequent ones were included. In addition, some characteristics of informal language including vowel lengthening which is converted to vowel repetition in writing for showing emphasis, surprise and other feelings were kept in informal sentences and edited in formal ones. As a matter of fact, it can be a shortcoming since we did not convey the feelings to the formal equivalents.

The last point is that Persian has two personal pronouns for singular address. It employs the second-person plural *šomâ* instead of the singular *to* as a sign of respect. A significant feature of colloquial Persian is a hybrid usage of the overt deferential second person pronoun and informal agreement forming a mismatch construction. It shows actually a different level of politeness (Nanbakhsh, 2011:1). In other words, plural pronoun with singular verb is used when the person being addressed is neither very intimate nor totally distant. A version of third person plural (*išun*) can be used in the same way. We kept this feature and did not change it in formal equivalents.

In the next section we are going to review the features of informal Persian and find out how users change the formal Persian in the informal writing. We are describing what we have seen in the data and explaining how we found the similar cases to develop a comprehensive corpus.

### 4.  Differences between Formal and Informal Persian

The level of informality varied among selected sentences. Some sentences only showed lexical changes. In example 1 every word of the sentence has different form in the formal equivalent.

(1) Informal: ye      hendune      vardâr!
              a       watermelon   take
    Formal:   yek     hendavâne    bardâr!
              a       watermelon   take
              (Take a watermelon!)

Some others underwent syntactic changes. Sentence 2 shows an example of word order change and preposition omission.

(2) Inf: diruz         bargašt- Ø         injâ.
       yesterday    came back-3rd SG     here
    F: diruz     be    injâ    bargašt- Ø.
       yesterday  to   here    came back-3rd SG
       (S/He came back here yesterday.)

Several other sentences had both kinds of changes. Many random differences including different kinds of abbreviations were only possible to be found by reading texts and other sources. On the other hand, there were changes that followed some morphological or phonological rules not necessarily regular led us to similar cases of the change. In order to examine each pattern, we searched it in general corpora including FarsNet (Persian wordnet) (Shamsfard, et al, 2010) and other online sources to find similar cases. Provided that the change had a reasonable frequency of occurrence, a few sentences from the sources were selected and recorded and in this way, tens or hundreds of instances of a change pattern were entered into the corpus. However, for the sake of space limits, only one example of each pattern is provided here. Next section will review the differences between formal and informal texts in four parts of phonological differences, morphological differences, syntactic differences and common mistakes.

### 4.1   Phonological Differences

There are many pronunciation distinctions between formal and informal Persian which have found their ways into written texts. Some are partly rule-based and follow the general rules of phonology and some others are users' creations. As mentioned earlier, language users sometimes break the rules of formal writing and devise different writing methods to be able to convey the tones and feelings along with the meaning. Some differences are as follows:

a. Many patterns of phonological reduction (mostly consonants) are observed in the informal Persian:
   (3) Inf: čan
       F: čan**d**
       (how many)

b. Sometimes speakers add a specified part to a formal word without adding any special meaning. These phonological additions, too, had some patterns to follow:
   (4) Inf: khâredj-**ak-**i
       F: khâredj-i
       (foreign)

c. Phonological alternation, being often rule-based, happen frequently in switching from formal to informal Persian:
   (5) Inf: âs**u**n
       F: âs**â**n
       (easy)

- d. Transposition of two adjoining sounds, known as adjacent metathesis, occurs normally in the informal Persian:
    - (6) Inf: qo**lf**
        F: qo**fl**
        (lock)

- e. There are some silent letters which do not correspond to any sound in the word's pronunciation. On the other hand, there are some sounds with no corresponding character in the word form. Since in some cases, the word forms follow the pronunciations in informal writing, people omit the silent letter or add the absent one:
    - (7) Inf: xâhar (خاهر)
        F: x**w**âhar (خواهر)
        (sister)

    "w" is silent in the formal word form. This change looks like writing the English word "enough" as "enaf".

- f. There are some Arabic phrases imported to Persian with their Arabic writing style (along with their articles and prepositions). Persian speaker usually changes their pronunciations and subsequently their word forms in the informal usage.
    - (8) Inf: išâllâ
        F: en-šâʔa-allâh
        (God willing)

- g. In order to break vowel sequences, People use different epenthetic consonants in informal speaking and subsequently in informal writing which may not match the usual epenthetic consonants:
    - (9) Inf: nobat-e    šomâ-**ʔ**-e
          turn-EZ[4]    you-**EPE**-is
        F: nobat-e  šomâ   ast.
           turn-EZ  you    is
           (It is your turn.)

- h. When words ending in /e/ are connected to words beginning with a vowel, both /e/ and the vowel are usually omitted in writing:
    - (10)   Inf: andâ**z-m**[5]
               size  my
           F: andâz**e-am**
               size  my
               (my size)

    Sometimes people omit only the second vowel (andâz**e-m**).

---

4 - Ezafe marker is placed into noun phrases, adjective phrases and some prepositional phrases linking the head and modifiers.
5 - Since short vowels do not appear in Persian writing, they are omitted in this example to show the change more clearly.

i. Some users, especially in social networks, deliberately change the letters of a word to emphasize something or ridicule or insult somebody:
   (11)   Inf: selebri**di**[6]
             F: selebri**ti**
                (celebrity)

## 4.2   Morphological Differences

A great deal of distinctions between formal and informal word forms can be studied in the field of language morphology. The morphological changes observed in this work are as follows:

a. The language users from younger generations are frequently observed to make up new infinitives from nouns:
   (12)   Inf: zang-idan
                call – infinitive suffix
             F: zang zadan
                call   hit
             (to telephone)

b. Some adverbs, conjunctions and question words can be used in plural forms in informal language:
   (13)   Inf: četori- y - **â** - st?
                how- EPE-PL-is
             F: četor   ast?
                how   is
             (How is it?)

c. In Persian, there is no number agreement between adjective and its modifying noun. In standard language, the plural suffix attaches to the noun while in informal Persian the plural suffix may be added to the adjective in a noun phrase:
   (14)   Inf: sib    qermez-**â**
                apple   red-PL
             F: sib - **hâ** - y - e    qermez
                apple-PL-EPE-EZ       red
                   (red apples)

d. In Persian script, some letters are written connected to their adjacent letter. When word forms are shortened in informal usage, they are sometimes written connected to each other and create new forms to process. For example, object marker râ changes into *ro* and *o* depending on the previous letter being a vowel or a consonant. Both *ro* and *o* may be written connected or unconnected:

---

6- offensive word

(15) Inf: man-o na-did- Ø
    me- obj marker    not-saw-3rd SG
    F: man   râ     na-did- Ø
    me    obj marker    not-saw-3rd SG
    (S/He did not see me.)

e. The shortened forms of many words have exactly the same forms; thus the ambiguity of informal writing is much more than formal writing. The data included the following examples:

- *ham* (also/too), *hastam* (am), and the first-person possessive pronoun are all shortened to "m":
    (16) Inf: mâmân-**m**
        mom-**m**
        (mom too/ I am a mom/ my mom)

- "i" can be a noun suffix, an indefinite article or second-person singular "to be" verb:
    (17) Inf: šâd-**i**
        happy-**i**
        (happiness / a happy [person] / you are happy)

- The informal form of *ast* (is) and the definite article have the same appearance (e):
    (18) Inf: ketâb-**e**
        book-**e**
        (it is a book/ the book)

- Informal object marker and the coordinating conjunction have a same form (o):
    (19) Inf: ketâb-**o**   bede   man.
        book-obj marker   give   me
        (give me the book)
    (20) Inf: ketâb-**o**  medâd
        book-and   pencil
        (book and pencil)

- Nunation or *tanvin* is an Arabic character appearing at the end of some Arabic loan words. It is written on "â" character, however, similar to short vowels, *tanvin* is usually omitted in writing. "â" is the shortened form of plural suffix, as well.
    (21) مثلا = for example
        مثلا = proverbs

A bigger number of examples were entered for ambiguous words in order for the machine to learn each meaning in different contexts.

f. Persian has two indefinite articles: *yek* and *i*. In informal Persian people normally use both together:

(22)   Inf: **ye**    doxtar-**i**
              One     girl-indef
         F: doxtar-**i**
              girl-indef
              (a girl)

g. Contrary to formal Persian, informal Persian has a definite article. Demonstratives were sometimes used in formal equivalents:

(23)   Inf: mard-**e**
              man-def
              (the man)
         F: **ân**    mard
              that    man
              (that man)

This article may also be used with adjectives. According to the context, the modified word was added in the formal equivalent:

(24)   Inf: qermez-**e**
              red-def
              (the red one)
         F: **ân**   [chiz]-e   qermez
              that   [sth]-EZ   red
              (that red [sth])

h. Clitics are vastly used in informal Persian. To come up with the formal equivalents, informal clitics were replaced by independent syntactic elements, as far as possible, in this study. However, there were informal clitics with no formal equivalents which needed to be omitted. The following examples show the cases of this change:

- Subject clitics on some third person intransitive verbs with no impact on meaning (25) and object clitics in clitic doubling structures (26):
    .
    (25)   Inf: sârâ       raft-Ø-**eš**.
                  sarah     went-3$^{rd}$ SG-sub clitic
             F: sârâ        raft-Ø.
                  sarah     went-3$^{rd}$ SG
                  (Sarah left.)

    (26)   Inf: sârâ       ro              did-am-**eš**
                  sarah     obj marker     saw-1$^{st}$ SG-obj clitic
             F: sârâ        râ              did-am
                  sarah     obj marker     saw-1$^{st}$ SG
                  (I saw Sarah.)

- Emphatic clitics:
    (27)   Inf: lebâs – â - t - o                         bešur – i – y -**â**
              Clothes-PL-your-obj marker            wash-2nd SG-EPE-clitic
           F: lebâs – hâ – y -at         râ              bešuy.
              clothes-PL-EPE-your    obj marker    wash
                     (Don't forget to wash your clothes.)

- In informal Persian some elements can be left-dislocated and left a clitic trace:
    (28)   Inf: sârâ       bâbâ-**š**         pir-e.
              sarah    dad-poss clitic    old-is
           F: bâbâ-ye      sârâ       pir   ast.
              dad-EZ       sarah      old   is
                     (Sarah's dad is old.)

## 4.3  Syntactic Differences

These kinds of changes were possible to be found only by searching in the sources. In other words, there was no specified pattern to follow. The changes observed in this study are listed below:

a. In general, Persian has a free word order, but there is a standard SOV order followed in formal language, while the informal does not often follow it. In this project, word order is standardized in the formal part of each sentence pair (22), except when an idiomatic meaning was intended (23):
   (29)   Inf: raft-am          madrese    man.
                went-1st SG        school       I
          F: man be    madrese       raft-am.
               I    to    school       went-1st SG
                     (I went to school.)

   (30)   Inf: boro  bâbâ! (idiom)
                 go     dad
          F: boro  bâbâ!
                (Go away!)

b. Omissions occur commonly in the informal language:

- The light verb in phrasal verbs can be omitted in informal Persian:
   (31)       Inf: bačče    qazâ      ro           **xorde**.
                    child    food   obj marker      eaten
              F: bačče     qazâ       râ         **xorde ast.**
                    child    food   obj marker      has eaten
                     (The child has eaten the food.)

- Omission of conjunctions, conditional elements and markers including *agar* (if), *vaqti* (when), *tâ* (so that), and *ke* (clause marker) is also common. The word order is sometimes affected by the omission:

    (32)   Inf: ber-i          madrese      bâsavâd       miš-i.
                go-2$^{nd}$SG    school       literate      become-2$^{nd}$SG
            F: **agar** be   madrese      berav-i,      bâsavâd       mišav-i.
                if    to    school       go-2$^{nd}$SG    literate      become-2$^{nd}$SG
                    (If you go to school, you will be literate.)

- Preposition stranding is disallowed in informal Persian, while a lot of preposition omission can be observed:
    (33)   Inf: raft-am         madrese.
                went-1$^{st}$SG    school
            F: be   madrese   raft-am.
                to    school     went-1$^{st}$SG
                    (I went to school.)

- The coordinating conjunction *va* (and) is sometimes omitted:

    (34)   Inf: qalam   kâqaz     biyâr.
                 pen     paper      bring
            F: qalam va  kâqaz     biyâvar.
                 pen  and  paper       bring
                    (Bring pen and paper.)

c. Simple past and present perfect have the same word form in informal written Persian (except for the 3$^{th}$ person singular).
   (35) Inf: xord-i
              ate-2$^{nd}$SG
         F: xord-i / xorde-ʔi
              ate-2$^{nd}$SG/ eaten-2$^{nd}$SG
              (ate/ have eaten)

d. A causal form of transitive verbs is sometimes used in informal Persian, which was turned to its non-causal transitive form in the formal member of sentence pairs:

   (36)   Inf: sârâ      šiša-ro              **šekund**- Ø.
                sarah    glass-obj marker     broke-3$^{rd}$ SG
           F: sârâ       šiše     râ           **šekast**- Ø.
                sarah    glass   obj marker    broke-3$^{rd}$ SG
                    (Sarah broke the glass.)

## 4.4 Common Mistakes

Common linguistic mistakes of users can again be syntactic, phonological or morphological. Mistakes were more observed in online comments and short messages. Similar to the two other changes, common mistakes could be traced by searching or following the patterns. Some of them are as follows:

a. Incorrect use of informal written form of copula *ast* (ه), Ezafe marker (ِ) and informal definite article (ه), all sounds like /e/, known as *Hekasre* error.

(37)    Inf: مامانه من (using article instead of Ezafe marker)
      my mom-def
    F: مامان من
      my mom-Ez
        (my mom)

b. Making plurals out of plural nouns
(38)    Inf: âqâ – y – **un** -**â**
      gentleman-EPE-PL-PL
    F: âqâ - y - **ân**
      gentleman-EPE-PL
        (gentlemen)

c. Adding Arabic *tanvin* (nunation) to Persian words:
(39)    Inf: telefon-**an**
       phone-tanvin
    F: telefon-**i**
      phone-noun suffix
       (by phone)

d. Using the verb *hast* (exists) instead of the copula *ast* (is).
(40)    Inf: in    xub    **hast**.
      this  good  exist
    F: in    xub    **ast**.
      this  good   is
     (This is good.)

e. Using a word mistakenly instead of another word with a similar pronunciation:
(41)    Inf: tas**f**iyehesâb
    F: tas**v**iyehesâb
      (settlement)

These kinds of mistakes which are much more common in informal writings, were tried to be covered in the database.

## 5. Results and Evaluation

The result of this research is available as a corpus of more than 50,000 pairs of formal-informal sentences along with a dictionary consisting formal-informal pairs of words and phrases. About half (49.77%) of the informal sentences needed syntactic changes besides lexical changes to be converted to formal ones, while the other half, could be converted just by changing the informal words. A detailed statistic is presented in table 2.

*Table 2. Statistics of the developed corpus*

| | |
|---|---|
| 50,014 | The number of input sentences |
| 12.32 | The average length of formal sentences |
| 11.36 | The average length of informal sentences |
| 529,286 | The number of word/phrase alignments |
| 71,842 | Number of unique word pairs (alignments) |
| 49.77% | The percentage of data with syntactic change |
| 49,397 | The dictionary size |

Raw data (informal sentences) is gathered from various sources. Table 3 shows the distribution of sentence sources in the final corpus. The row 'Myself" means that the sentence is not extracted from a source and is rather generated by the linguists.

*Table 3. Distribution of different sources in the final data*

| source | # of sentences |
|---|---|
| Web | 26,014 |
| Twitter | 5,308 |
| Instagram | 4,747 |
| Myself | 3,528 |
| Movie (including movies, dramas and movie subtitles) | 3,282 |
| Messenger | 2,751 |
| Weblog | 2,400 |
| Book | 1,984 |
| total | 50,014 |

For extrinsic evaluation of the corpus, we used it in a deep model of an informal to formal converter and compared the results with a rule-based method. Experiments show that using a deep Bert2Bert architecture (Falak Aflaki, 2021) which is trained on our corpus leads to bleu score of 81.69% on sentences with 15-25 words length, while the rule-based method gains 79.74% bleu score on the same test set.

## 6. Conclusion and further work

This study was conducted to develop an informal-formal language corpus for Persian language for the purpose of natural language processing. In order to achieve this aim, many available sources of informal writing were explored to recognize its particular features and build a well-organized and operative dataset.

The minimum possible changes such as transpositions, additions and omissions were applied to make the formal equivalents in order not to change the original meaning, however, there are evidently shortcomings such as omitting some informal segments of emphasis and feelings in formal equivalents which led to omit a part of meaning that was inevitable according to our instructions. This issue can be addressed in future studies.

Moreover, although we tried to cover the differences between informal and formal Persian writing as far as possible, there are certainly cases we have missed.

## Acknowledgement

This project is developed under the grant no. 11/61955 from Vice-Presidency for Science and Technology of I.R. Iran. Authors would like to thank the members of corpus development team especially Ms. Elham Fekri, Ms. Parastoo Falak Aflaki and Mr. Mostafa Karimi Manesh for their great contributions.

## Statement of Conflicting Interests
The authors state that there is no conflict of interest.

## References

Aftekhari, S. A. 2005. Language Economy and Its Influence on Colloquial Persian. Literary Research, Volume 2, Issue 7: 25-60. Tehran, Iran. [In Persian]

Armin, N., & Shamsfard, M. 2011. Transforming Persian Informal Texts Using N-gram. Paper presented at the 16th CSI Computer Conference, Tehran, Iran.

Falak Aflaki, P. 2021. *Informal to Formal Persian Transformation.* (Bsc), Shahid Beheshti University, Tehran, Iran.

Kabiri, R., Karimi, S., & Surdeanu, M. 2022. Informal Persian Universal Dependency Treebank. (https://arxiv.org/abs/2201.03679) (Accessed 2022-05-20.)

Khojasteh, H. A., Ansari, E., & Bohlouli, M. 2020. 'LSCP: Enhanced Large Scale Colloquial Persian Language Understanding'. (https://arxiv.org/abs/2003.06499) (Accessed 2022-05-20.)

Naemi, A., Mansourvar, M., Naemi, M., Damirchilu, B., Ebrahimi, A., & Kock Wiil, U. 2021. Informal-to-Formal Word Conversion for Persian Language Using Natural Language Processing Techniques. Paper presented at the 2nd International Conference on Computing, Networks and Internet of Things.

Nanbakhsh, G. 2011. *Persian Address Pronouns and Politeness in Interaction.* (Doctoral Dissertation), University of Edinburgh, Edinburgh. Retrieved from https://era.ed.ac.uk/bitstream/handle/1842/6206/Nanbakhsh2011.pdf?sequence=2&isAllowe (Accessed 2022-05-20.)

Rasooli, M. S., Bakhtyari, F., Shafiei, F., Ravanbakhsh, M., & Callison-Burch, C. 2020. 'Automatic Standardization of Colloquial Persian'. (https://arxiv.org/abs/2012.05879v1) (Accessed 2022-05-20.)

Shamsfard, M., Hesabi, A., Fadaei, H., Mansoory, N., Famian, A., Bagherbeigi, S., Assi, S. M. 2010. Semi-Automatic Development of Farsnet; the Persian Wordnet. Paper presented at the Proceedings of 5th Global WordNet Conference, Mumbai, India.

Tabibzadeh, O. 2020. Orthography of Colloquial Persian: Based on Works of Fiction and Drama Spanning a Century (1918-2028). *Institute for Humanities and Cultural Studies*, Tehran, Iran. [In Persian]

Tatrar, A. 2018. Patterns of Persian writing in messengers (A Case Study of Telegram Messenger). *Special issue of Farhangestan (Dastour)*. Number 14. Tehran, Iran. [In Persian]